\documentclass[twoside,12pt]{article}

\usepackage{amssymb, amsmath}
\usepackage{jmlr2e}
\usepackage{subfigure}
\usepackage{algorithm2e}
\ShortHeadings{Technical Report}{Technical Report}
\firstpageno{1}

\begin{document}

\title{Graph-sensitive Indices for Comparing Clusterings}

\author{\name Zaeem Hussain \\
\addr Department of Applied Mathematics \\
University of Washington\\
\email zaeem.hussain@gmail.com
\AND
\name Marina Meil\u{a}  \\
       \addr Department of Statistics\\
       University of Washington\\
       Seattle, WA 98195-4322, USA\\
       \email mmp@stat.washington.edu}

\maketitle

\begin{abstract}%   <- trailing '%' for backward compatibility of .sty file
This report discusses two new indices for comparing clusterings of a set of points. The motivation for looking at new ways for comparing clusterings stems from the fact that the existing clustering indices are based on set cardinality alone and do not consider the positions of data points. The new indices, namely, the Random Walk index (RWI) and Variation of Information with Neighbors (VIN), are both inspired by the clustering metric Variation of Information (VI). VI possesses some interesting theoretical properties which are also desirable in a metric for comparing clusterings. We define our indices and discuss some of their explored properties which appear relevant for a clustering index. We also include the results of these indices on clusterings of some example data sets.
\end{abstract}

%\begin{keywords}
%  Bayesian Networks, Mixture Models, Chow-Liu Trees
%\end{keywords}

\section{Introduction}

\begin{figure}
	\centering
	\subfigure[]{\includegraphics[width=1.3in]{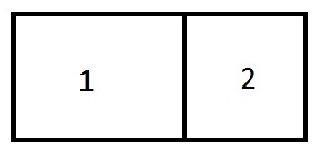}}
          \subfigure[]{\includegraphics[width=1.3in]{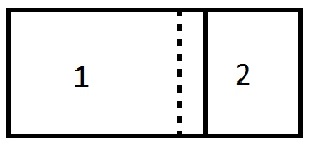}}
          \subfigure[]{\includegraphics[width=1.3in]{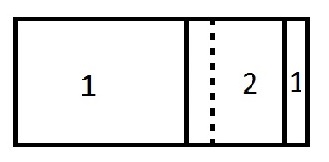}}
	\caption{3 different clusterings of the same data. In (b), some points from group 2 in (a) are put in group 1. In (c), the same number of points that changed labels in (b) again change from label 2 to 1, but the points that change are different from the points that changed in (b).}
	\label{fig:data}
\end{figure}

The problem of comparing clusterings essentially seeks to quantify the similarity or dissimilarity between two clusterings or partition of a dataset. This comparison can be needed in a variety of cases. For example, suppose we have a desired or correct clustering of a dataset  and an algorithm that also outputs a clustering of the same data. An index that compares two clusterings would then be required to see whether the output of the algorithm is close to the correct solution or not. Such an index would also be required when we have the results of two such algorithms and want to decide which algorithm outputs a solution closer to the correct clustering. This is just one of many cases where indices for comparing clusterings are needed. \\
Clusterings can be compared based on different properties and there are multiple indices that are developed by focusing on those different properties. However, all clustering comparison criteria are based on the \emph{Confusion matrix} or \emph{contingency table} of the clusterings being compared [Meil\u{a}, 2007]. Formally, let the clusterings being compared be denoted by $C=\{C_1,C_2,...C_K\}$ and $C'=\{C'_1,C'_2,...C'_{K'}\}$ and the number of clusters in both be $K$ and $K'$ respectively. $C_1,...C_K$ are mutually disjoint subsets and so are the clusters in $C'$. The \emph{confusion matrix} is a $K\times K'$ matrix whose $kk'$-th element is the number of points in the intersection of the clusters $C_k$ and $C'_{k'}$, where $k \in \{1,2,...K\}$ and $k' \in \{1,2,...k'\}$. Since all clustering indices can be defined using the confusion matrix, they are based on the counts of clusters and their intersections alone and ignore any other relationships the points may have. In other words, when comparing two clusterings, these indices depend only on the number of points that go into each cluster and not on the distances between points in a cluster and in different clusters. Consider, for example, the different clusterings of a dataset in Fig \ref{fig:data}. Any existing index if calculated between the first two clusterings would yield the same answer as that of the index calculated between the first and last clustering. However, as can be seen clearly from the figure, the second and third clusterings are not the same and should not be judged to be at the same 'distance' from the first partitioning. The problem, then, is to define an index with some desirable properties that would compare two clusterings while also incorporating spatial correlations between points in the dataset, if there are any.\\

\section{Variation of Information}
The Variation of Information (VI) is a clustering metric that also suffers from the above mentioned limitation in comparing clusterings. However, it possesses some interesting theoretical properties which are desirable in any index for comparing clusterings. VI is calculated by assigning to each cluster a probability that a point picked at random would fall into that cluster. Precisely, let the total number of points in the dataset be $n$. Call one clustering $C$ with a total of $K$ clusters and the other clustering $C^{'}$ with $K^{'}$ clusters. Denote the number of points in cluster $C_k$ in $C$ with $n_k$ and the number of points in $C^{'}_{k^{'}}$ in $C^{'}$ be $n_{k^{'}}$. The probability that a point chosen at random would fall in cluster $C_k$ is
\begin{equation}P(k)=\frac{n_k}{n}\end{equation}
and similarly the joint probability that a point chosen at random falls in cluster $C_k$ in $C$ and in cluster $C^{'}_{k^{'}}$ in $C^{'}$ is
\begin{equation}P(k,k^{'})=\frac{|C_{k} \cap C^{'}_{k^{'}}|}{n}\end{equation}
This probability can be calculated using the confusion matrix defined in the previous section. Let the $kk'$-th element of the confusion matrix be denoted by $n_{kk'}$. Based on the definition of the confusion matrix, $n_{kk'}=|C_{k} \cap C^{'}_{k^{'}}|$ and so the probability can be calculated as
\begin{equation}P(k,k^{'})=\frac{n_{kk'}}{n}\end{equation}
Using these probabilities, we can define the entropy for clustering $C$ as
\begin{equation}H(C)=-\sum_{k=1}^{K}P(k)\mbox{log}P(k)\end{equation}
and similarly for $C'$. The mutual information between $C$ and $C'$ is defined as
\begin{equation}I(C,C')=\sum_{k=1}^{K}\sum_{k'=1}^{K'}P(k,k')\mbox{log}\frac{P(k,k')}{P(k)P(k')}\end{equation}
Using these quantities, the variation of information between two clusterings is then defined as
\begin{equation}VI(C,C')=H(C)+H(C)-2I(C,C')\end{equation}
which using simple arithmetic on entropies, can be shown to be of the form
\begin{equation}\label{VI}VI(C,C')=H(C|C')+H(C'|C)\end{equation}
As mentioned before, (VI) possesses certain desirable properties. Some important properties are that it is bounded and that it is a metric, thus introducing a notion of distance on the space of clusterings. Some of its theoretical properties will be discussed in detail in the subsequent sections as they are compared with the properties of our proposed indices.

\section{Proposed Indices}
A natural way to view the points in a dataset with distance defined between any pair of points is as a weighted undirected graph. Such a graph can be described by the $n\times n$ \emph{similarity matrix} $S$ whose $ij$-th element represents the similarity or the edge weight between points $i$ and $j$. A special case of this would be a graph with no weights on the edges which may be represented by the binary adjacency matrix $A$ where the $ij$-th entry is 1 if there is an edge between point $i$ and $j$ and 0 otherwise. Hence, with the points represented as nodes in a weighted undirected graph, one may define indices which also look at the labels of the neighbors of a point instead of comparing clusterings by just comparing labels of each point independently of its neighbors.

\subsection{Index based on Markov Random walks}
Based on the idea of considering the data as a graph, we can define an index using the random walks view of the graph. First, let us review some concepts on transition probabilities using the Markov Random walk theory [Meil\u{a} and Shi, 2001]. The similarities between points of the graph be in $S$ are nonnegative where $S_{ij} = S_{ji} \geq 0$ represents the similarity between points $i$ and $j$ of the graph. The degree of a node $i$ is then defined as $d_i=\sum_{j=1}^{n}S_{ij}$ and matrix $D$ is a diagonal matrix formed with the degrees of the nodes. The stochastic matrix $T$ is obtained by "normalizing" the similarity matrix as
\begin{equation}T=D^{-1}S\end{equation}
The entry $T_{ij}$ of $T$ represents the probability of going from node $i$ to $j$ given that we are in node $i$. The $stationary$ $distribution$ of the Markov chain, denoted $\pi_{i}^{\infty}$, is defined as
\begin{equation}\pi_{i}^{\infty}=\frac{d_i}{\sum_{i=1}^{n}d_i}\end{equation}
Let $k_t$ represent the label of the point traversed at the current time step $t$ during the random walk in the first clustering and $k'_t$ represent its label in the second clustering. Therefore, $k_t \in \{1,2,...,K\}$ and $k'_t \in \{1,2,...,K'\}$.
Now, assuming the Random Walk starts in the stationary distribution, the probability of going from a point in cluster $C_k$ to a point in $C_l$ in one step given we are in cluster $C_l$ at previous time $t-1$ can be denoted as $P_{C_kC_l}=Pr(C_k \rightarrow C_l|C_k)=Pr(k_{t-1}=C_k, k_t=C_l|k_{t-1}=C_k)$ and defined as
\begin{equation}P_{C_kC_l}=\frac{\sum_{i\in C_k,j\in C_l}\pi_{i}^{\infty}T_{ij}}{\pi^{\infty}(C_k)}=\frac{\sum_{i\in C_k}\pi_{i}^{\infty}\sum_{j\in C_l}T_{ij}}{\pi^{\infty}(C_k)}\end{equation}
Similarly, the probability $Pr(k_t|k'_t,k_{t-1})$ can be obtained by
\begin{equation}\begin{array}{lcl}Pr(k_t=C_l|k'_t=C'_m,k_{t-1}=C_k)&=&Pr(C_k\rightarrow C_l,C_k\rightarrow C'_m |C_k\rightarrow C'_m)\\&=&\frac{Pr(C_k\rightarrow (C_l\cap C'_m))}{Pr(C_k\rightarrow C'_m)}\end{array}\end{equation}
Calculating the transition probabilities this way allows us to condition the label of a point in one clustering on its label in the other clustering as well as on the label of its neighbor. Using these probabilities we can define the first of our indices, the Random Walk index (RWI), as
\begin{equation}\label{rwiIndex}RWI(C,C')=H(k_t|k'_t,k_{t-1})+H(k'_t|k_t,k'_{t-1})\end{equation}
This index sums up the uncertainty in the label of a point given the label of the point visited in the previous step and the label of the same point in the other clustering. By including information about the label of the just traversed point, we are adding the label information of one neighbor of each point. Comparing equation \eqref{rwiIndex} with \eqref{VI} we can see that the difference in the 2 indices is the inclusion of label of a neighbor in the calculation of \eqref{rwiIndex} whereas VI does not incorporate any such information. Since we are adding another conditioning variable in the entropy terms, this index will always be smaller than VI. One point to note is that the time step $t$ does not matter since we just need transition probabilities for one step and so this $t$ does not figure in the calculations of this index.\\
Since the number of probability entries that will be calculated is $O(K^3)$ where $K$ is the maximum number of clusters in one of the two clusterings, the running time for the algorithm to calculate this index is cubic in the number of clusters present in either of the two clusterings. So, if the number of clusters in the first clustering is $K$ and in the second clustering is $K'$, the running time to calculate this index would be $O(KK'(K+K')N)$ where $N$ is the total number of points in the data set.
\subsection{Variation of Information with Neighbors}
Another index we propose as a comparison metric for clusterings also takes into account the information about labels of neighbors of a point. However, rather than quantifying how much information the labels of neighbors give about the point, this metric measures the amount of information the labels of a neighborhood of points in one clustering give about the labels of the same neighborhood in the other clustering. Formally, the metric is defined as
\begin{equation}VIN(C,C')=H(N(X)|N(Y))+H(N(Y)|N(X))\end{equation}
where again $C$ and $C'$ are the two clusterings being compared. $X$ is the label of a point in the first clustering and $N(X)$ represents the labels of the neighbors of that point, and the point itself, in the first clustering. Similarly, $Y$ is the label of the same point in the second clustering and $N(Y)$ represents the labels of this point and its neighbors in the second labeling.\\
Whereas the previous proposed index was based on probabilities calculated from the similarity matrix of the graph using Markov random walk theory, the probabilities here are simply based on counting the different kinds of neighborhoods in the clustering. In other words, for a set of $n$ points, there are $n$ neighborhoods, one for each point. Each neighborhood is characterized by the label of its 'central' point and the numbers of this point's neighbors that belong to each of the clusters in the labeling. Two neighborhoods are considered equal, or belonging to the same class, when the following three conditions are met:
\begin{enumerate}
\item The labels of both 'central' points are same.
\item The numbers of neighboring points of each of the 'central' points are equal.
\item The number of neighboring points belonging to a cluster should be the same in both the neighborhoods.
\end{enumerate}
As an example, consider a graph of 100 nodes where each node is given one of 3 different labels: $a$, $b$ and $c$. One node $i$ is labeled $a$ and has 5 neighbors, 2 of which are labeled $a$, the other two are labeled $b$ and one is labeled $c$. Another node $j$ is also labeled $a$ and has 5 neighbors with the same number of nodes in each label as neighbors of $i$. So the neighborhoods of $i$ and $j$ would be regarded as equal or belonging to the same category. On the other hand, a node $k$ which also has 5 neighbors with the same distribution of labels but itself labeled $c$ would fall into a different category.\\
Using this classification of neighborhoods, we can calculate the the conditional probabilities $P(N(X)|N(Y))$ by counting the number of neighborhoods belonging to a category in the second labeling and then looking at the categories of those neighborhoods in the first clustering. More precisely, to calculate $P(N(X)=S_X|N(Y)=S_Y)$, where $S_X$ and $S_Y$ are sets of point labels in the first and second clustering respectively, we first count the number of points whose neighborhoods in the second clustering are equal to $S_Y$ in the sense described above. From those points, the number of points whose neighborhoods in the first clustering equal $S_X$ is noted. The probability then is just a fraction of these two numbers.\\
Following is the algorithm to compute VIN:\\
\\
\begin{algorithm}[H]
 \SetAlgoLined
 \KwData{Row vectors $u$ and $v$ of size $1\times n$ representing labels of points in $C$ and $C'$, $n\times n$ adjacency matrix $A$}
 \KwResult{Variation of Information with Neighbors}
 \%Stack both row vectors vertically\\
 $U \leftarrow $repmat($u$, $n$, 1);\\
 $V \leftarrow $repmat($v$, $n$, 1);\\
 \%Compute element by element product\\
 $A_u \leftarrow U.*A$;\\
 $A_v \leftarrow V.*A$;\\
 $B_u \leftarrow$ sortrwd($A_u$);\\
 $B_v \leftarrow$ sortrwd($A_v$);\\
 $u' \leftarrow $ finduniquerows($B_u$);\\
 $v' \leftarrow $ finduniquerows($B_v$);\\
 $VIN \leftarrow$ VI($u'$,$v'$);\\
 \caption{Algorithm to compute VIN}
\end{algorithm}

The inputs to the function to compute this index would contain two arrays of same size containing point labels in both the clusterings and the $N\times N$ adjacency matrix for the graph. The adjacency matrix can be used to obtain two $n\times n$ matrices $A_u$ and $A_v$ whose rows represent the labels of the neighborhoods of the points. The entry $A_u(i,j)$ would equal the label of point $j$ in the first clustering if it is connected to $i$. If $i$ and $j$ are not connected, $A_u(i,j)=0$. The diagonal entry $A_u(i,i)$ would represent the label of point $i$ in the first clustering. $A_v$ would similarly represent the labels of points in the second clustering. Now, if the neighborhoods of two points in the first clustering are equal, it would mean that the diagonal entries of their corresponding rows would be same and the remaining elements of one row would be a permutation of the remaining elements in the second row. Hence, comparing of neighborhoods can be done by comparing the rows of each these matrices. The \emph{sortrwd} operation in the algorithm first shifts the diagonal entry of each row in the matrix to the beginning of the row and sorts the rest of the row for all rows in the matrix. The operation \emph{finduniquerows} indexes the unique rows in the matrix such that same rows get the same label and returns labels for the rows. Finally, the function \emph{VI} simply calculates the variation of information between two clusterings represented by vectors in its input.\\
If the maximum degree of any node in the graph is $K$, the maximum number of non zero elements in any row would be $K+1$. Finding the different kinds of neighborhoods, then, could be done by first sorting the rows, which would be done in $O(KlogK)$ for each row, and comparing the rows with each other. Again, since the maximum length of any row is $K+1$, the row comparisons can be done in $O(NK)$ using a Radix tree [Morrison, 1968]. Thus, the running time of such an algorithm would be $O(NKlogK)$.
\section{Some properties of RWI and VIN}
\begin{figure}
	\centering
		\includegraphics[width=0.8\textwidth]{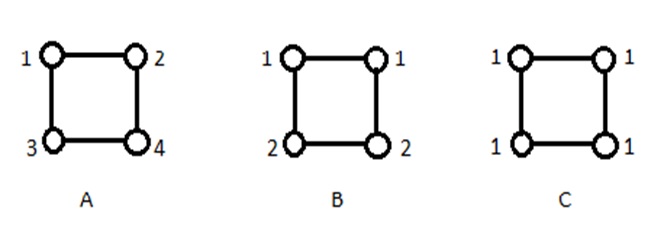}
	\caption{Three different clusterings of a set of 4 points. The number with each node represents its label.}
	\label{fig:tring_ineq}
\end{figure}

As mentioned before for the Variation of Information, a desirable property of a clustering comparison index is that it should be a metric. As can be seen from the definition of the first proposed index in Equation \eqref{rwiIndex}, the index is symmetric as well as nonnegative, being zero only for identical clusterings. Thus, if this index satisfies the triangle inequality, then it will be a metric and will give a measure of the closeness of two clusterings while also taking into account the associations between data points. However, the proposed Random Walk Index was computed for a simple example of 4 points as shown in Figure \ref{fig:tring_ineq} and upon calculation it was found that
\begin{equation}RWI(A,B)+RWI(B,C)<RWI(A,C)\end{equation}
which implies that the Random walk index is not a metric on the space of clusterings.\\
The other proposed index, Variation of Information with Neighbors, however, satisfies the triangle inequality for arbitrary clusterings $A$, $B$ and $C$ of a set of points.
\begin{equation}VIN(A,B)+VIN(B,C)\geq VIN(A,C)\end{equation}
This index satisfies the other 2 properties (nonnegativity and symmetry) and is also zero for identical clusterings. Thus, it is a metric on the space of clusterings. It also satisfies the following relation for any clustering $C$
\begin{equation}VIN(\hat{1},C)+VIN(C,\hat{0})=VIN(\hat{1},{0})\end{equation}
where $\hat{1}$ is the labeling in which each point of the data set is given a different label and $\hat{0}$ is the other extreme, where each point is given the same label. This relation is also satisfied by VI but not by the Random Walk index. The VI with neighbors can be considered to be a more general case of VI in the sense that if the graph is completely connected or not connected at all, VIN as defined before reduces to VI. RWI also reduces to VI when the graph is not connected at all, but now when the graph is completely connected.\\
There is an interesting property that holds true for the Variation of Information and which makes it a more intuitive distance over the space of clusterings. If a clustering $C'$ is obtained from $C$ by splitting $C_k$ in a number of clusters, VI between $C$ and $C'$ is equal to the probability of cluster $C_k$ times the entropy of the clusters obtained from $C_k$. Formally, assume $C'$ is obtained from $C$ by splitting $C_k$ into clusters $C'_{k_1},...C'_{k_m}$. The cluster probabilities are
\begin{equation}P'(k')=\begin{cases} P(k'), & \mbox{if } C'_{k'}\in C \\ P(k'|k)P(k), & \mbox{if } C'_{k'}\subseteq C_k\in C \end{cases}\end{equation}
where $P(k'|k)$ for $k'\in \{k_1,...k_m\}$ is
\begin{equation}P(k_l|k)=\frac{|C'_{k_l}|}{|C_k|}\end{equation}
and its entropy, which represents the uncertainty associated with splitting $C_k$, is
\begin{equation}H_{|k}=-\sum P(k_l|k)\mbox{log}P(k_l|k)\end{equation}
Then [Meil\u{a} 2007],
\begin{equation}VI(C,C')=P(k)H_{|k}\end{equation}
This property also induces the additivity of composition over VI, which says that if two clusterings are obtained by further segmenting the same clustering, VI between the two clusterings is a weighted sum of VI between the partitions of each cluster in the bigger clustering. \\
The VI with Neighbors holds a weak form of additivity of composition. To see this, we first observe that the VIN also basically computes the VI between two partitions. However, these partitions are obtained from the original clusterings being compared and are essentially \emph{refinements} of the original clusterings. A \emph{refinement} $D$ of a clustering $C$ is a partitioning which preserves the boundaries in $C$ but some of the clusters in $C$ are further split. The reason why the clusterings compared for VI with Neighbors are refinements of the original clusterings is that each point is relabeled based on the labels of its neighbors and no two points that have different labels in the original labeling are reclassified as belonging to the same category when the neighborhoods are compared. Hence, VI with Neighbors can be written as
\begin{equation}VIN(C,C')=VI(D,D')\end{equation}
where $D$ and $D'$ are refinements of $C$ and $C'$ respectively and are based on the neighborhoods of the points. In VI, the additivity of composition holds because the distance between two clusterings depends only on the clusters that vary between the two partitions. However, for VI with neighbors, the distance is computed between clusterings which are refinements of the originals and these refinements are obtained by looking at the directly connected neighbors of the points. So even if two clusterings differ only in one cluster, their refinements as dictated by VIN will be influenced by points outside that cluster which are directly connected to points inside it. Hence, VIN can be considered as satisfying weak additivity of composition in the sense that the relation upon splitting a cluster as for VI above holds for VIN only when the cluster that is split is not connected to the rest of the points in the graph. If the points in that cluster have edges with other points in the graph, such a relation does not hold for VIN.
\section{Experiments and Results}
Since the indices in this report were proposed to address the limitation of VI that it does not include the neighborhood of data points, the examples of clusterings considered here will be those on which VI does not provide a satisfactory answer. The three kinds of graph that will be considered are: chain with evenly spaced points, Gaussian data with similarities between pairs of points encoded by weights on edges, and images, where similarities between pixels depend on their spatial distances from each other. In all the examples we will have three clusterings where the last two are different modifications of the first clustering and are judged by VI to be at the same distance from the first one.
\subsection{One dimensional Chain}
\begin{figure}
	\centering
		\includegraphics[width=0.8\textwidth]{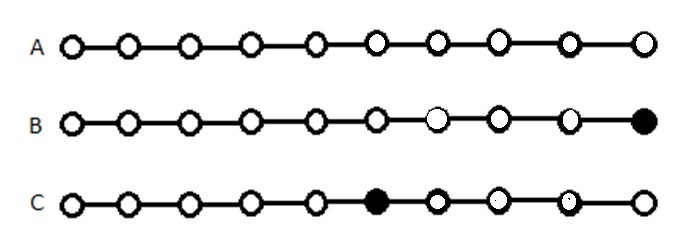}
	\caption{Three different clusterings of a set of 10 points}
	\label{fig:example2}
\end{figure}

\begin{figure}
	\centering
		\includegraphics[width=0.8\textwidth]{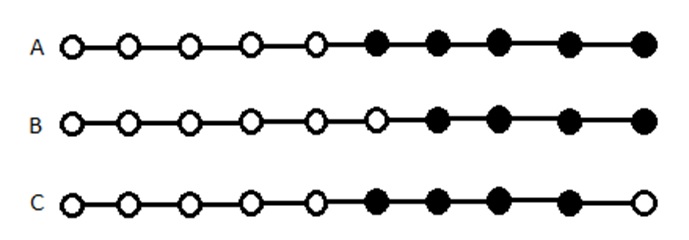}
	\caption{Three different clusterings of a set of 10 points}
	\label{fig:example1}
\end{figure}

The first case we consider is that of a chain with all the points belonging to the same cluster. Two new clusterings are obtained from this clustering by relabeling one of the points of the chain to a different label. In one clustering, a point from the middle is chosen and in the second, a point at one of the ends is chosen for relabeling. An instance of this case is shown in Figure \ref{fig:example2} where the total number of points is 10. For the random walk index, all the weights on the edges are uniformly set to 1. Intuitively, we would expect the clustering $C$ to be closer to $A$ than $B$. This is indeed what is observed with with both the proposed indices. The values of random walk index and VI with neighbors between $A$ and $B$ are greater than the corresponding values of these indices between $A$ and $C$. It must be remembered that VI would judge both $B$ and $C$ to be at the same distance from $A$.\\
Next we test the two proposed indices on a scenario based on the same situation as shown in Figure \ref{fig:data}. In this case, the original labeling has two clusters where one half of the line belongs to one cluster and the other half belongs to another cluster. Two new clusterings are obtained from this by taking a certain number of points from one cluster and relabeling them as belonging to the other cluster. One clustering is obtained this way by relabeling the points closest to the boundary between the two clusters and the other is obtained by relabeling the points at the end of the line. An example of this again with 10 points is shown in Figure \ref{fig:example1}. Again the Variation of Information will judge the second clustering to be at the same distance from first one as the third clustering, although based on the location of the points, the second and third clusterings should be at different distances from the first one. This is indeed what is observed when the two proposed indices are used to compare these clustering. However, the results seem contrary to intuition because both the indices judge the third clustering to be closer to the first one than the second clustering. The Random walk index was computed based on two different similarity matrices. The first one only included information about the two adjoining neighbors of each point in the chain. The other similarity matrix used included edges between all the points in the chain, with the edge weights inversely related to the distance between the points. Still, in both cases, the index judged the third clustering closer to the first segmentation than the second one.\\
\subsection{Gaussian data}
\begin{figure}
	\centering
		\includegraphics[width=0.8\textwidth]{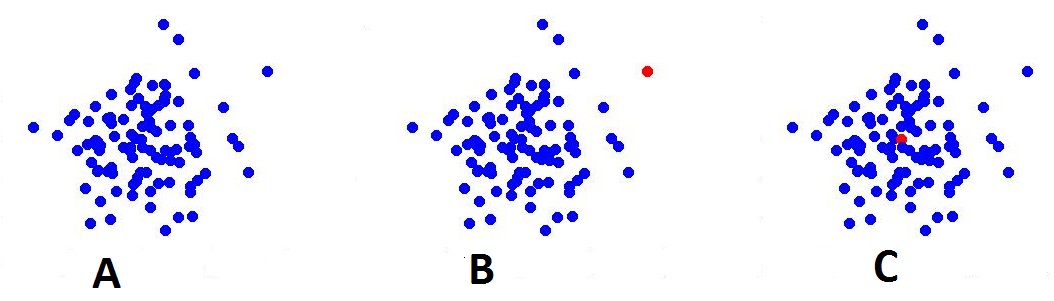}
	\caption{Three different clusterings of a set of 100 points obtained from a 2D Gaussian distribution}
	\label{fig:gaussian}
\end{figure}

Next we consider Gaussian data where the edge weights between points represent the similarity between them. If the distance between points $i$ and $j$ is $d_{ij}$, the similarity, $s_{ij}$, between the points is calculated as $s_{ij}=e^{-d_{ij}^2}$. For VI with neighbors, the adjacency matrix is obtained by setting a threshold where all edges with weights above the threshold are kept and the rest are dropped. If the weights are obtained from spatial distances between the points, this is equivalent to taking an $\epsilon$ neighborhood around each point. The clusterings we use here are similar to the first case of the chain example discussed in the previous section. Initially, all the points have the same label. One clustering is then generated by relabeling the point that is farthest from the mean and the other clustering is generated by relabeling the point closest to the mean. An example for the two dimensional Gaussian data with 100 points is shown in Figure \ref{fig:gaussian}.\\
A total of 100 simulations of this scenario were run with the covariances between the $x$ and $y$ coordinates set to 0 and variances for both $x$ and $y$ set to 1. In the majority of the simulations, both the random walk index and VI with neighbors judge the clustering where the farthest point from the mean is relabeled to be closer to the original clustering than the one where the point closest to the mean is relabeled. The results are summarized in table below. The 2 middle columns list the means of the corresponding indices computed first on clusterings $A$ and $B$ and then on clusterings $A$ and $C$. The last column lists the number of mistakes made by the indices from the 100 trials. In the context of the example in figure \ref{fig:gaussian}, the Random Walk Index judged $B$ to be closer to $A$ than $C$ 96 times while VI with neighborhoods did not make a single mistake in the 100 simulations in declaring $B$ to be closer to $A$ than $C$.
\begin{center}
  \begin{tabular}{| l | c | c | c | }
  \hline
    Indices & Mean of d(A,B) & Mean of d(A,C) & Errors \\ \hline
    VI & 0.0243 & 0.0243 & 100 \\ \hline
    RWI & 0.0355 & 0.1362 & 4 \\ \hline
    VIN & 0.0142 & 0.0782 & 0 \\ \hline
  \end{tabular}
\end{center}
\subsection{Images}
\begin{figure}
	\centering
		\includegraphics[width=0.3\textwidth]{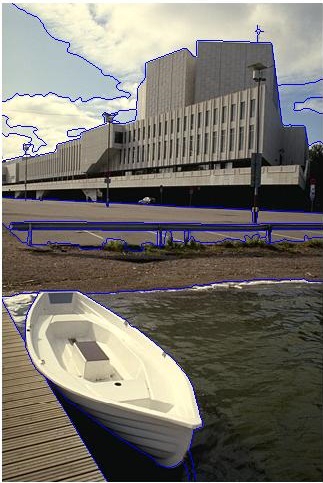}
	\caption{An image with true segmentation boundaries}
	\label{fig:picture}
\end{figure}

\begin{figure}
	\centering
		\includegraphics[width=0.8\textwidth]{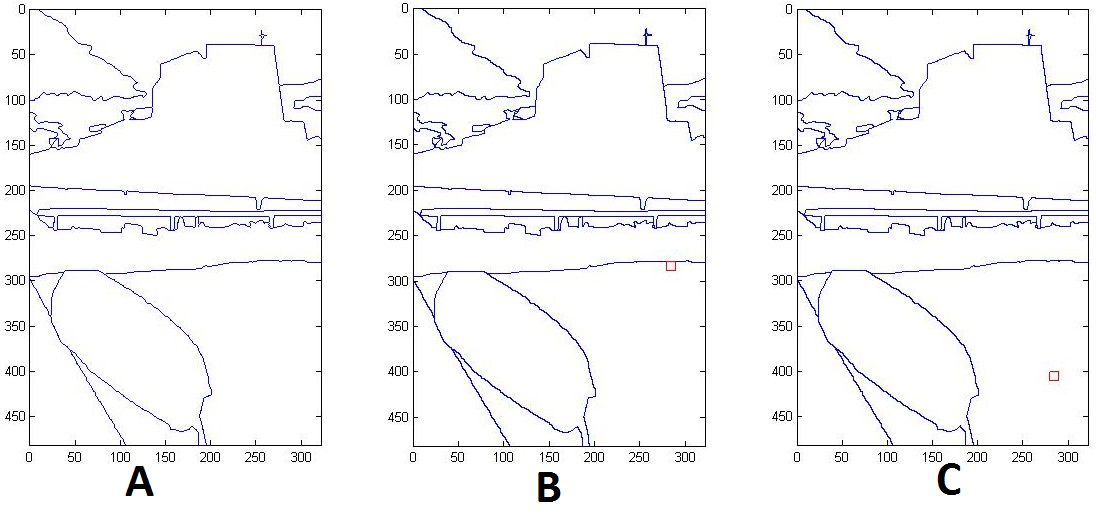}
	\caption{True segmentation of the image and two perturbations}
	\label{fig:image_square}
\end{figure}

\begin{figure}
	\centering
		\includegraphics[width=0.8\textwidth]{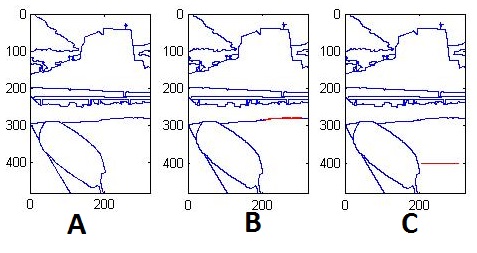}
	\caption{True segmentation of the image and two perturbations}
	\label{fig:image_line}
\end{figure}

We compute the indices between segmentations of the image in figure \ref{fig:picture} which also shows the true boundaries of the image. The edge weights between pixels for the random walk index are negative exponentials of the square of the spatial distances between them. For simplicity and memory constraints, only a $5\times 5$ neighborhood around each point is considered. For VI with neighbors, all the points in the $5\times 5$ neighborhood of a pixel are considered neighbors of the pixel.\\
We create clusterings from the original by relabeling 100 pixels of water in the image as belonging to the cluster for the dirt in the image. Two cases of such perturbations of the original segmentation are considered. The first one, shown in figure \ref{fig:image_square}, has the clusterings formed by relabeling a $10\times 10$ square of the pixels whose boundaries are shown in red. Both the indices, however, judged $C$ to be closer to $A$ than $B$. The results for this example are tabulated as follows:
\begin{center}
  \begin{tabular}{| l | c | c | }
  \hline
    Indices & d(A,B) & d(A,C) \\ \hline
    RWI & 0.0184 & 0.0172 \\ \hline
    VIN & 0.0064 & 0.0062 \\ \hline
  \end{tabular}
\end{center}
The second case that is considered is shown in figure \ref{fig:image_line}. Here, instead of relabeling a $10\times 10$ square of pixels, the 100 pixels just along the boundary between land and water are relabeled in $B$ and a horizontal line of 100 pixels, which is quite far from the original 'dirt' segment pixels, is relabeled as belonging to 'dirt' cluster on land to obtain the perturbed clustering $C$. For this example, VI with neighbors judged $B$ to be closer to $A$ than $C$. However, the random walk index again judged $C$ to be closer to $A$ than $B$. The results are in the following table:
\begin{center}
  \begin{tabular}{| l | c | c | }
  \hline
    Indices & d(A,B) & d(A,C) \\ \hline
    RWI & 0.0335 & 0.0248 \\ \hline
    VIN & 0.0087 & 0.0094 \\ \hline
  \end{tabular}
\end{center}
\section{Conclusion}
We presented two criteria for comparing clusterings of a data set that take into account the similarity between the points. The first index, which we call the Random Walk index, is based on probabilities calculated from the similarity matrix of the graph using Markov Random Walk theory. The second proposed index, the Variation of Information with neighbors, counts different neighborhoods based on the labels of the points in the neighborhoods. Both these criteria were tested on examples to see whether they possess the properties desirable in a clustering comparison criterion.\\
The Random Walk index was found not to satisfy the triangle inequality and is thus not a metric on the space of clusterings. VI with neighbors, however, can be shown to satisfy the triangle inequality and other conditions for a distance and so is a metric on the space of clusterings. Both the indices were observed to judge clusterings differently based on the location of the points in the data. Some of the results agreed with what one might expect how the distance between a clustering should behave but the results on other example apparently indicate otherwise. Further exploration in this avenue with more examples might give a clearer picture of how these indices might fit with human intuition about the similarity between clusterings of a set of points. The property of splitting a cluster, where the comparison metric should not depend on points whose labels remain the same over the clusterings being compared, was also checked for the proposed indices. The Random Walk index did not always satisfy this property but was rather observed to depend on the similarity matrix. VI with neighbors, on the other hand, satisfies a weak form of this property where the distance between two clusterings depends on the points directly connected to the cluster that varies even if the labels of those points remain unchanged across the clusterings being compared.\\
This report is meant as an introductory document on the ideas for comparing clusterings of a set of points while incorporating the information about the distances between points. Some of the properties that are theoretically interesting for a comparison index were checked on some basic examples, but a detailed analysis with larger examples would be required to establish these indices as standard comparison criteria. There are other interesting theoretical avenues for exploration with these indices as well, such as the comparison with the meet of the clusterings and identifying nearest neighbors of the clusterings according to these indices.
\section{Appendix}
Proofs to follow.
\section{Referecnes}
\begin{enumerate}
\item Meil\u{a}, Marina. "Comparing clusterings— an information based distance." Journal of Multivariate Analysis 98.5 (2007): 873-895.
\item Meil\u{a}, Marina, and Jianbo Shi. "A random walks view of spectral segmentation." (2001).
\item Morrison, Donald R. "PATRICIA—practical algorithm to retrieve information coded in alphanumeric." Journal of the ACM (JACM) 15.4 (1968): 514-534.
\end{enumerate}
\end{document}